\RequirePackage{fix-cm}
\documentclass[twocolumn]{svjour3}          
\smartqed  
\usepackage{graphicx}
\usepackage[pdftex]{color}
\usepackage[font=footnotesize,labelfont=bf]{subcaption}
\usepackage{multirow}
\usepackage{array}
\usepackage{longtable}
\usepackage{tikz}
\usepackage{array}
\usepackage{booktabs}
\usetikzlibrary{positioning}
\usepackage{amssymb, mathtools}
\DeclareMathOperator*{\argmin}{arg\,min}
\DeclareMathSizes{5}{5}{4}{4}
\usepackage{siunitx}

\begin{document}

\title{Deep Multi-Facial Patches Aggregation Network For Facial Expression Recognition
}


\author{ Ahmed Rachid Hazourli  \and
         Amine Djeghri         \and
         Hanan Salam   \and
         Alice Othmani 
}


\institute{ Ahmed Rachid Hazourli \at
              Universit\'e Paris Saclay, 91400 Orsay, France \\
           \and
           Amine Djeghri \at
              Sorbonne Universit\'e, 75006 Paris, France \\
         \and Hanan Salam \at 
         Emlyon,  69130 Écully, France \\
        \and Alice Othmani  \at
         Univ Paris Est Creteil, LISSI, F-94400 Vitry, France \\
         Corresponding author: \email{alice.othmani@u-pec.fr} 
}

\date{Received: date / Accepted: date}

\maketitle



\begin{abstract}


In this paper, we propose an approach for Facial Expressions Recognition (FER) based on a deep multi-facial patches aggregation network. Deep features are learned from facial patches using deep sub-networks and aggregated within one deep architecture for expression classification . 
Several problems may affect the performance of deep-learning based FER approaches, in particular, the small size of existing FER datasets which might not be sufficient to train large deep learning networks. Moreover, it is extremely time-consuming to collect and annotate a large number of facial images. To account for this, we propose two data augmentation techniques for facial expression generation to expand FER labeled training datasets.
We evaluate the proposed framework on three FER datasets. Results show that the proposed approach achieves state-of-art FER deep learning approaches performance when the model is trained and tested on images from the same dataset. Moreover, the proposed data augmentation techniques improve the expression recognition rate, and thus can be a solution for training deep learning FER models using small datasets. The accuracy degrades significantly when testing for dataset bias.



\end{abstract}

\graphicspath{{Images/}}

\section{Introduction}

Human face conveys a significant amount  of information about identity, gender, ethnicity, age and emotion. Through facial expressions, the face is able to communicate countless emotions without any spoken words. Moreover, facial expressions provide substantial evidence of human's level of interest, understanding, mental state \cite{el2005real}, as well as a continuous feedback of agreement or disagreement within a social interaction. 
According to the universality hypothesis of facial expressions of emotion established since Darwin's seminal work \cite{JackFacial2012}, all humans communicate six basic internal emotional states (happy, surprise, fear, disgust, anger, and sad) using the same facial movements. On the other, research in psychology has shown that facial expressions is person-dependent and the intensity of expressions varies highly among different individuals and cultures \cite{marsh2003nonverbal}. 

Emotion recognition from face images has attracted considerable attention in a wide range of Human Computer Interaction (HCI) applications such as robotics \cite{zhang2020capsule}, education \cite{tongucc2020automatic}, biometrics \cite{rida2020motion} and health informatics \cite{simcock2020associations}.
Giving computers the capacity to infer human users' affective states is key for a successful natural human-computer interaction within these applications.

The rise of deep learning techniques had led researchers to explore different deep architectures and methods for FER \cite{LiDeep2018} in the past decade. Deep neural networks have significantly improved the performance of FER tasks compared to traditional approaches \cite{khorrami2015deep,mollahosseini2016going,minaee2019deep} due to their capacity to automatically learn both low and high level descriptors from facial images without the need of human intervention.

Despite considerable advancements in the field, automated Facial Expression Recognition (FER) has remained a challenging task. As a matter of fact, most existing methods still lack generalizability when applied to images captured in the wild.
This is mainly due to: 
(1) Variations in spontaneous expressions manifestations among individuals, as well as non-controlled environmental conditions such as illumination, occlusions and non-frontal head poses. 
(2) Deep learning approaches are highly dependent on the availability of huge labeled facial expressions datasets. 
(3) Deep networks have huge number of parameters (several millions) to learn which increases the risk of model over-fitting.

On the other hand, psychological studies have shown that the most contributing facial regions in conveying facial expressions are the nose, mouth and eyes \cite{CohnComputerized1995}. 
Previous studies have applied patch-based approaches for FER systems \cite{Zhong2012,Nicolle2015,liu2013aware,du2020self}.
These approaches typically divide the face into different patches and extract appearance descriptors on these patches which are then combined and passed to a classifier to predict the facial expression.
Other approaches relied on the detection of Action Units which are then translated to the corresponding facial expressions based on the FACS \cite{FACS2002}. 
Although proved efficient, such approaches still use hand-crafted appearance-based features to encode the appearance of the facial expressions. Moreover, the extracted features are directly aggregated with no further strategy to learn higher level representation which may be essential for mapping AU to facial expressions compared to deep learning based FER approaches. 

In the context of image quality, style, aesthetic recognition \cite{lu2015deep} proposed a deep multi-patch aggregation network training approach to overcome the shortening of deep randomly selected single-patch approaches conventionally used for these applications. Within their approach, the input image is represented with a small set of randomly selected patches which are associated with the image’s training label. Patches are aggregated through two statistics and sorting network layers and fed to multiple shared columns in the neural network.

Inspired by this work and building up on previous patch-based and deep learning-based FER approaches, we propose a Multi-Facial Patches Aggregation Convolutional Neural Network (MFP-CNN) for expression classification from face images. Facial landmarks are detected and used to generate seven local patches of the most contributing regions for facial expressions (eyes, eyebrows, nose, mouth).  Facial regions patches are then fed into seven deep sub-networks (Multi-Patches CNN) to learn the variations of these regions. Outputs of sub-networks are aggregated and fed to a Convolutional Neural Networks (CNN) which learns mapping these patches to facial expressions. 

Acquiring rich facial expressions datasets can be problematic since labeling emotions is extremely hard, time consuming and requires several experts interventions, which in turn can suffer from low inter-rater agreement due to annotators subjectivity (different labels by different annotators for the same expression/emotion). This rises the need to develop new data augmentation techniques that enable researchers to fully benefit from the advances in deep learning architectures. 

To overcome the lack of labelled datasets required to train Deep Neural Networks, to improve also the performance of the proposed Multi-Facial patches-based CNN and to avoid overfitting, we propose two data augmentation techniques to increase and to diversify the training data and then expand the labeled training. A Conditional Generative Adversarial Network (cGAN) and Transformation Functions are proposed for Facial Expression Generation.  

We evaluate our deep architecture on three cross-domain FER datasets. As a matter of fact, evaluating the performance of cross-domain FER systems is a challenging topic. For instance, by studying crossing facial expressions and ethnicity using artificial intelligence on big data face images, researchers can solve the longest debates in the biological and social sciences of the universality of facial expressions emotions \cite{JackFacial2012}.  

An overview of the rest of the paper is as follows: in Section~\ref{related_work} the literature of FER is reviewed. Section~\ref{proposed_method} describes the proposed method. Section~\ref{results} present quantitative results. A conclusion and future perspective works are given in Section~\ref{conclusion}.


\begin{table*}[]
\begin{tabular}{p{3.0cm}|p{2.0cm}|p{2.5cm}|p{2.5cm}|p{1.5cm} |p{2.0cm}}
\toprule
Dataset & Challenge & Samples & Nb subjects/Faces & Nb expressions & lab-controlled
     \\ \midrule
AFEW 7.0 \cite{Dhall2017} & EmotiW 2017 & 1809 videos & - & 7 & NO
 \\ \midrule
SFEW \cite{DhallVideo2015} & EmotiW 2015 & 1766 images & - & 7 & NO
 \\ \midrule
HAPPEI \cite{DhallAutomatic2015}& - & 4886 & 8500 & 6 & NO
 \\ \midrule
GENKI \cite{Whitehill2009} & - & 63,000 images & approximately as many different human subjects & 3 & NO 
  \\ \midrule
GEMEP-FERA \cite{ValstarFirst2011} & FERA Emotion sub-ch & 289 videos & 13 sub & 5 & NO 
  \\ \midrule
 MMI \cite{PanticWeb2005} & - & 740 images + 848 videos & 19 & AUs & YES 
  \\ \midrule
 FER2013 \cite{Goodfellow2013} & ICML 2013 & 35887 images & - & 7 & NO 
  \\ \midrule
 CK+ \cite{Lucey2010} & - & 593 sequences & 123 & 7 & YES 
 \\ \midrule
 JAFFE \cite{Lyons1998} & - &  213 & 10 & 7 & YES 
  \\ \midrule
 Oulu-CASIA \cite{ZhaoFacial2011} & - &  2880 videos & 80 & 6 & YES 
   \\ \midrule
 MultiPIE \cite{GrossMulti2010}  & - &  750000 & 337 & 6 & YES \\ 

\bottomrule
\end{tabular}
\caption{An overview of facial expression datasets.
}
\label{datasets_overview}
\end{table*}


\begin{table*}[]
\centering
\begin{tabular}{p{3.0cm}|p{6.5cm}|p{4.0cm}|p{1.5cm}}
\toprule
Method & Approach & Dataset & Accuracy 
  \\ \midrule
  \multirow{3}{*}{Zhang et al. (2017) \cite{Zhang2017}} & Deep evolutional spatial-temporal networks (PHRNN-MSCNN) & CK+ & 98.5\% \\
 & & OULU-CASIA & 86.25\% \\
  & & MMI & 81.18\% 
  \\ \midrule
 \multirow{3}{*}{Sun et al. (2017) \cite{Sun2019}} & Multi-channel Deep spatial-Temporal feature fusion neural network (MDSTFN) & CK+ & 98.38\% \\
 & & RaFD & 99.17\% \\
  & & MMI & 99.59\% 
  \\ \midrule
  \multirow{4}{*}{Kuo et al. (2018) \cite{Kuo2018}} & frame-based and frame-to-sequence FER framework & CK+ (frame-based) & 97.37\% \\
  & & CK+ (frame-to-seq) & 98.47\% \\
  & & Oulu-Casia (frame-based) & 88.75\% \\
  & & Oulu-Casia (frame-to-seq) & 91.67\% 
  \\ \midrule
   \multirow{3}{*}{kim et al. (2017) \cite{Kim2017}} & deep generative-contrastive networks
 & CK+ & 97.93\% \\
 & & Oulu-Casia & 86.11\% \\
  & & MMI & 81.53\% 
    \\ \midrule
  \multirow{4}{*}{Yang et al. (2018) \cite{YangIdentity2018}} & Identity-Adaptive Generation (IA-gen)
 & CK+  & 96.57\% \\
  & & OULU-CASIA & 88.92\% \\
  & & BU-3DFE & 76.83\% \\
  & & BU-4DFE & 89.55\% 
  \\ \midrule
   \multirow{3}{*}{Jung et al. (2015) \cite{Jung2015}} & Deep Temporal Appearance-Geometry Network (DTAGN) & CK+ & 97.25\% \\
   & & OULU-CASIA & 81.46\% \\
   & & MMI & 70.24\% 
   \\ \midrule
   \multirow{2}{*}{Yu et al. (2018) \cite{Yu2018}}  & Deeper Cascaded Peak-piloted Network (DCPN) & CK+ & 98.60\% \\
   & & OULU-CASIA & 86.23\% 
    \\ \midrule
    \multirow{2}{*}{Zhao et al. (2016) \cite{Zhao2016}} & Peak-piloted deep network (PPDN) & CK+ & 97.3\% \\
    & & OULU-CACIA & 72.4 \% \\

\bottomrule
\end{tabular}
\caption{An overview of deep learning-based approaches for FER.}
\label{overview_deep_fer}
\end{table*}

\section{Related work}
\label{related_work}

Since the early 1970s, the universality and cultural differences in facial expressions of emotions have been studied \cite{EkmanUniversals1987}. A strong evidence of cross-cultural agreement in the judgment and the universality of facial expressions has  shown that 'universal facial expressions' fall into seven categories: 
happiness, sadness, anger, fear, surprise, and disgust. 
Accordingly, Ekman \textit{et al}. \cite{FACS2002} proposed the Facial Action Coding System (FACS) which encodes facial movements in terms of atomic facial actions called Action Units (AUs) that can be measured and used for the recognition of affective states and emotions.  Ekman's work inspired many researchers to develop automatic multi-modal facial expressions recognition approaches. In the following, existing approaches for FER as well as most popular FER datasets are presented. 

\subsection{Facial Expression Recognition approaches}
Several works have been proposed for facial emotions classification from static images in the literature \cite{DhallStatic2011,ValstarFirst2011,DhallEmotion2011,IoannouEmotion2005,DahmaneEmotion2011,Hernandez2007,Tong2007,Zhong2012,Nicolle2015,Yan2012,KimDeep2013,Mollahosseini2016,PoriaConvolutional2016,Yu2015,LiDeep2018,NgDeep2015} or from continuous video input \cite{Bartlett2003,Cohen2003,Michel2003}. 


For instance, a system that automatically finds faces in the visual video  stream and codes facial expression dynamics in real time with seven dimensions (neutral, anger, disgust, fear, joy, sadness, surprise) is proposed in \cite{Bartlett2003}. 
To demonstrate the potential of the proposed system, a real time `emotion mirror' is developed to render a 3D character that mimics the emotional expression of a person in real time . 
Another system for classification of facial expressions from continuous video input is proposed in \cite{Cohen2003} to learn from both labeled and unlabeled data. 

Common approaches in facial expression recognition constitute two main steps: (1) features extraction and (2) classification. Low and high level features are first extracted and then fed into a classifier to output the predicted expression label.
Feature extraction methods in FER approaches can be generally categorized into three categories: hand-crafted,  metric-learning based and deep-learning based. \\

\textbf{Hand-crafted features based:} Such methods design and extract descriptive features in an attempt to capture the appearance and shape of facial expressions. Various hand-crafted feature schemes have been proposed for facial emotion classification. These include: action units (AUs) \cite{Tong2007}, Local Gabor Binary Pattern histograms combined with Active Appearance Models coefﬁcients \cite{senechal2012facial}, Gabor magnitude representation using a bank of Gabor filters \cite{Bartlett2003}, histogram of gradients (HOG) \cite{DahmaneEmotion2011,Bartlett2003}, Local Phase Quantisation (LPQ) features for encoding the shape and appearance information \cite{DhallEmotion2011}, Facial Animation Parameters (FAPs) \cite{IoannouEmotion2005}, motion features \cite{Cohen2003,Michel2003} and texture descriptors like Gray Level Co-occurrence Matrix \cite{Hernandez2007}.  Although proved efficient, using hand-crafted appearance-based features to encode the appearance of facial expressions is limiting  and requires human intervention to design find the best features. Moreover, the extracted features are directly aggregated with no further strategy to learn higher level representation which may be essential to correctly map facial Actions to facial expressions.\\

\textbf{ Metric learning based :} a two-stage multi-task sparse learning (MTSL) framework is proposed to efficiently locate discriminative patches \cite{Zhong2012}. Two multi-task extensions of Metric learning for kernel regression (MLKR) for facial action unit intensity prediction was proposed in \cite{Nicolle2015}. The Hard Multi-task regularization for MKLR introduces a more constrained common representation between tasks than a standard multi-task regularization. 
An adaptative discriminative metric learning (ADML) is also proposed for facial expression recognition by imposing large penalties on inter-class samples with small differences and small penalties on inter-class samples with large differences simultaneously \cite{Yan2012}. The interest of such approaches lies in their capacity to discover underlying common structure between tasks. \\ 

\textbf{Deep learning based :} recently, high-level semantic features were designed based on deep neural networks architectures for facial emotions recognition. The multi-level neural networks perform a series of transformations on the face image. On each transformation, a denser representation of the face is learnt. Abstract features are learnt in the deeper layers which allow a better prediction of emotions. 
A comparison between a suite of Deep Belief Network models (DBN) and baseline models shows improvement in audio-visual emotion classification performance when using DBN models. Further, the three-layer DBN outperformed the two-layer DBN models \cite{KimDeep2013}. In \cite{Mollahosseini2016}, the approach aims to go deeper using deep neural networks, a network of two convolutional layers each followed by max pooling and then four inception layers is proposed and shows an improvement of performance on seven publicly available facial expression databases. Liu  \cite{liu2020facial} combines a semi-supervised deep learning model for feature extraction and a regularized sparse representation classifier.  

Different deep networks architectures are proposed for facial emotion learning. Among them, those which study the \textit{spatial and the temporal information} and incorporate them into different networks.
Zhang \textit{et al.} \cite{zhang2020facial} combine a double-channel weighted mixture deep convolution neural networks (WMDCNN) based on static images and deep CNN long short-term memory networks of double-channel weighted mixture(WMCNN-LSTM) based on image sequences. WMDCNN network recognizes facial expressions and provide static image features for WMCNN-LSTM network which utilizes the static image features to further acquire the temporal features of image sequence for accurate recognition of facial expressions. 
Deep evolutional spatial-temporal networks (PHRNN - MSCNN) is proposed in \cite{Zhang2017}. The method is a fusion of temporal network for modelling dynamical evolution called Part-based Hierarchical Bidirectional Recurrent Neural Network (PHRNN) and a spatial network for global static features called Multi-signal Convolutional Neural Network. In the PHRNN, facial landmarks are divided into four parts based on facial physical structure and fed to the BRNN models, while the MSCNN takes pairs of frames as input with both recognition and verification signals and feeds the frame to a CNN. 
Similar to the PHRNN-MSCNN, the Deep Temporal Appearance-Geometry Network (DTAGN) \cite{Jung2015} combines deep temporal geometry network (DTGN) which extracts temporal geometry features from temporal facial landmark points and deep temporal appearance network (DTAN) which extracts temporal appearance features from image sequences. 
Another Multi-channel Deep spatial-Temporal feature fusion neural network (MDSTFN) is proposed \cite{Sun2019} where the temporal information is defined as the optical flow from the peak expression face image and the neutral face image, while the spatial information is defined by the gray-level image of emotional face. 

Other deep networks consider a \textit{peak-piloted feature transformation} by using the peak expression images to supervise the intermediate responses of non-peak expression images \cite{Zhao2016,Yu2018}. The \textit{Conditional Generative Adversarial Network (CGAN)} \cite{Mirza2014} are also used to generate basic facial expressions in the Identity-Adaptive Generation (IA-gen) network \cite{YangIdentity2018}. Then, features are extracted using the pre-trained CNN and the query image is labeled as one of the six basic expressions based on a minimum distance in feature space.  More complex and original architecture is proposed in \cite{Kim2017} through a deep generative-contrastive networks. It is a combination of a generative model, a contrastive model and a discriminative model. The contrastive representation is calculated at the embedding layer of deep network by comparing a given (query) image with the reference image.  Other approaches combine data from \textit{different modalities} for emotion recognition and sentiment analysis. For instance, in \cite{PoriaConvolutional2016}, the convolutional recurrent multiple kernel learning model combines features extracted from video, speech and text. A convolutional RNN is used for extracting features from video data. A survey on deep FER including datasets and algorithms is given in \cite{LiDeep2018} and an overview of deep learning-based approaches for FER is given in Table~\ref{overview_deep_fer}.
 
\subsection{FER Datatsets}
Traditionally, emotion recognition databases are acquired in laboratory-controlled conditions where subjects pose in a particular emotion. The face images are generally taken with static backgrounds, controlled and invariant illumination and without any movement \cite{PanticWeb2005,GrossMulti2010,ZhaoFacial2011,Lyons1998,Lucey2010}.
Emotion recognition from facial images in wild conditions is a relatively new and challenging problem in face analysis. Several datasets have been proposed in recent years  where photos of the subjects are acquired in real-world scenarios without any lab-controlled conditions \cite{Goodfellow2013,Dhall2017,DhallVideo2015,DhallAutomatic2015,Whitehill2009,ValstarFirst2011}. 
To be able to cope with such challenging conditions, FER methods should be robust to indoor, outdoor, different color backgrounds, occlusions, background clutter and face misalignment. Happy People Images (HAPPEI) \cite{DhallAutomatic2015}, Acted Facial Expressions in the Wild (AFEW) \cite{DhallCollecting2012}, Static Facial Expressions In the Wild (SFEW) \cite{DhallStatic2011} and GENKI \cite{Whitehill2009} are recent databases for emotion recognition in the wild with real-world scenarios. The existing databases for emotion recognition presents mainly universal emotions like angry, disgust, fear, happy, neutral, sad and surprise. Recent databases use continuous labelling in the arousal and valence scales or an increased number of emotion dimension. For instance, \cite{wang2020learning} proposed the Fine-grained Facial Expression Database (F2ED) with 54 expression emotions, such as calm, embarrassed, pride, tension and so on, including abundant emotions with subtle changes. An overview of facial expression datasets is given in Table~\ref{datasets_overview}. \\

 




\begin{figure*}[]
  \centering \includegraphics[width=.9\textwidth]{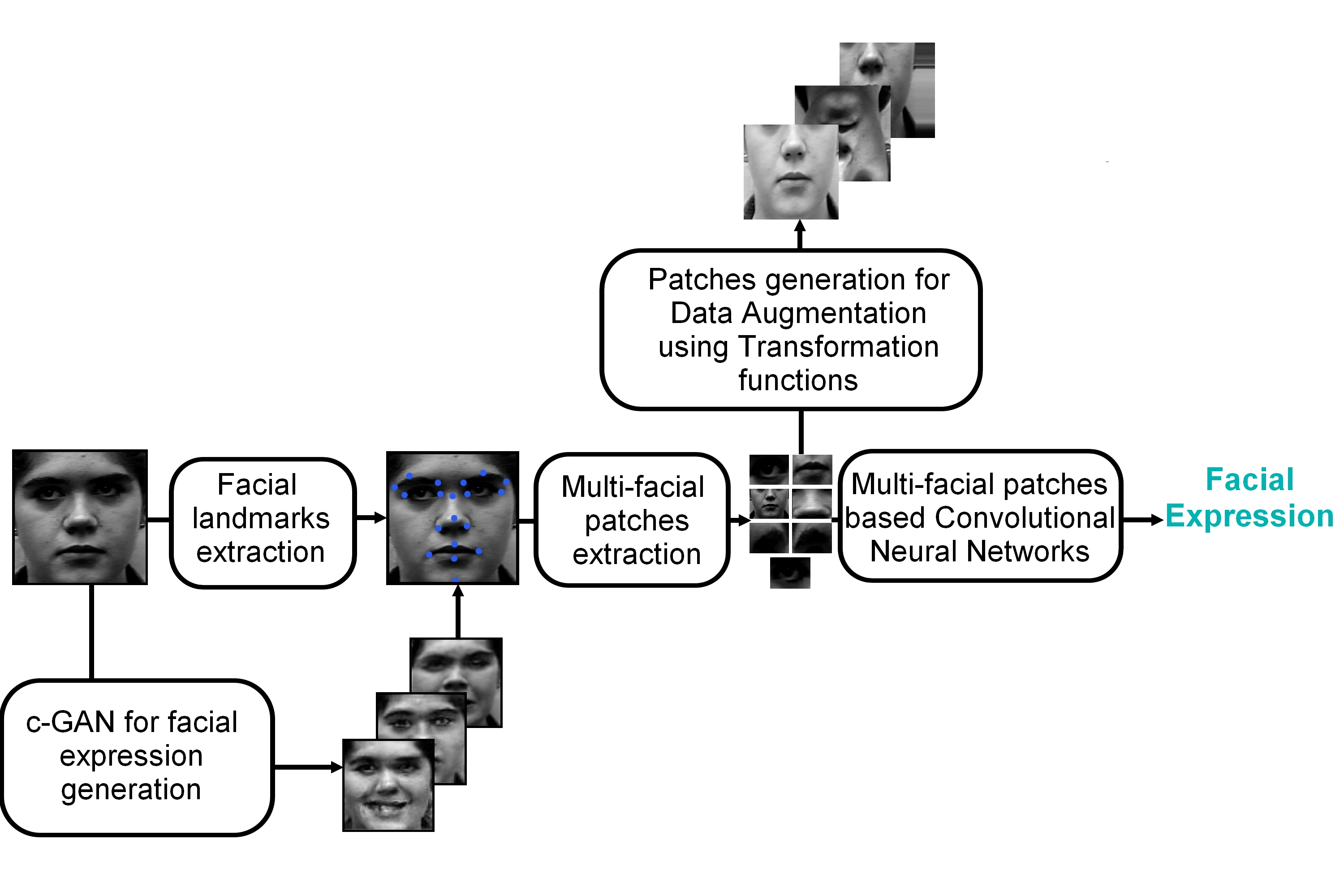}
  \caption{Illustration of the proposed approach.}
  \label{diagram_proposed_approach}
\end{figure*}

\section{Proposed Method}
\label{proposed_method}

Our approach is divided into 5 steps. 
(1) Given a face image, facial landmarks are first detected. 
(2) Facial patches are then cropped around facial landmarks (section~\ref{multi_patches_extraction}) and (3) fed to a multi-patches based Convolutional Neural Networks (section~\ref{multi_patches_based_cnn}). 
(4) The response of the sub-networks of all patches are aggregated at two dense layers to classify the facial expression (section~\ref{multi_patches_based_cnn}). Figure~\ref{diagram_proposed_approach} illustrates the proposed framework.\\

A large amount of labeled data is required to train deep neural networks. However, emotion recognition datasets are relatively small and do not have a sufficient quantity of data. Moroever, FER datasets contain very often data with imbalanced classes. The annotation task for facial expression generally requires the intervention of experts. Moreover, the labeling procedure is extremely time-consuming, difficult and prone to subjective errors. To bypass the need for a large dataset for deep models training, we propose and evaluate two data augmentation techniques based on cGAN \cite{Mirza2014} and simple transformation functions to generate facial expressions from neutral images.  This is explained in detail in sections~\ref{cGAN} and ~\ref{sec_patches_generation}.

\subsection{Multi-facial patches Extraction}
\label{multi_patches_extraction}

\begin{figure*}[]
  \centering \includegraphics[width=\textwidth]{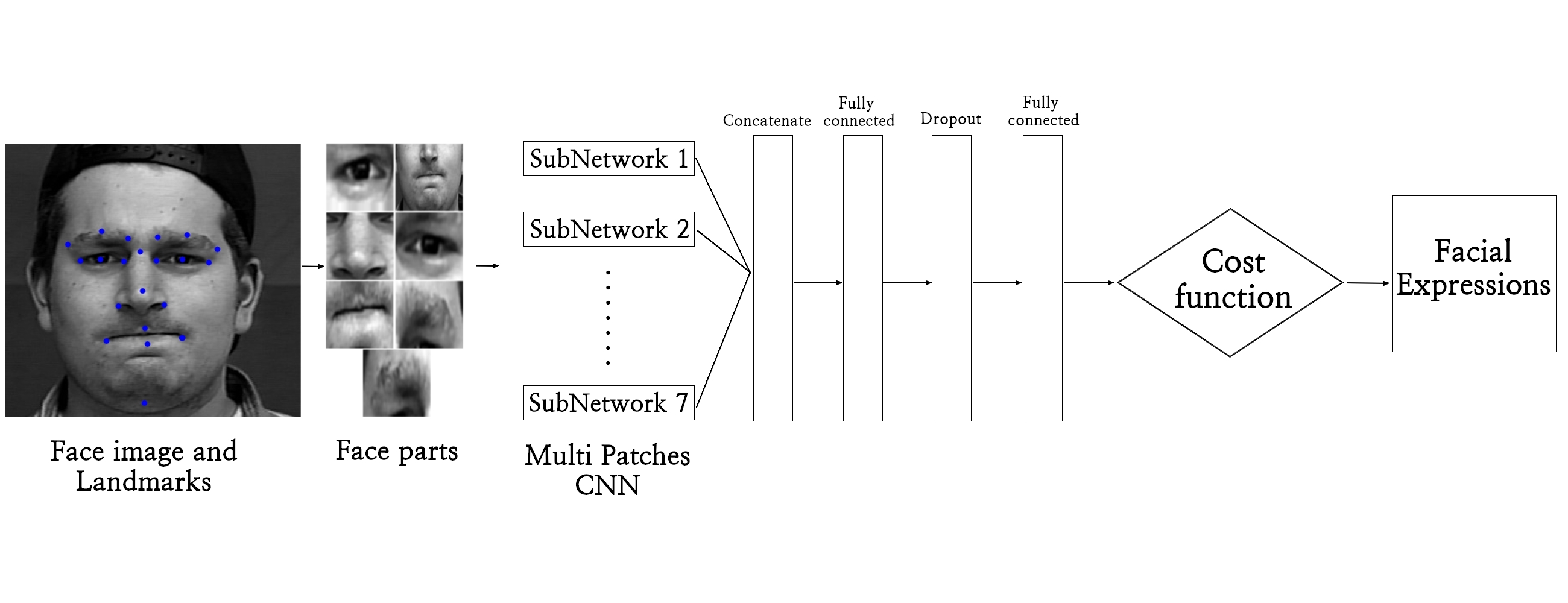}
  \caption{Multi-Facial Patches Aggregation Convolutional Neural Networks (MFP-CNN). For each image, the face is detected, aligned and  facial landmarks are extracted. Facial patches are extracted around facial landmarks and each is fed to a sub-network. The structure of each sub-network is shown in Fig.~\ref{sub-network}}
  \label{MFP-CNN}
\end{figure*}

Facial expression is presented by the dynamic variation of key parts of the human face (e.g. eyes, nose and mouth). Our approach makes use of the variation of local parts which are learned and fused together within a deep architecture to obtain the variation of the whole face.

The first step of the proposed framework concerns the extraction of local aligned facial patches. Facial landmarks are used to localize and to represent salient regions of the face (eyes, eyebrows, nose, mouth and jawline). We use the facial landmark detector of Kazemi et al., 2014 \cite{Kazemi2014}. In this approach, a fully discriminative model based on a cascade of boosted decision forests to regress the position of landmarks from a sparse set of pixel intensities is performed. This method has proved to provide accurate landmarks detection in the majority of cases.

After detecting the facial landmarks, we perform an alignment step using the positions of the eyes. Alignment is simply a transformation from an input coordinate space to an output coordinate space so that all faces are centered, eyes lie on a horizontal line, and faces are scaled such that the faces sizes are nearly identical. 
Facial landmarks proved to have better performance for face alignment than Haar cascades or HOG detectors since the bounding box provided less precision to estimate the eye location  in the latter as compared to landmarks indexes. 


Seven patches are extracted around the facial landmarks corresponding to face key parts which represent the left and right eyes, the nose, the mouth, the left and right eyebrows, and the jaw. The extracted facial patches allow the study and comparison of their dynamic variation corresponding to each facial expression.  

\begin{figure*}[]
  \centering \includegraphics[width=\textwidth]{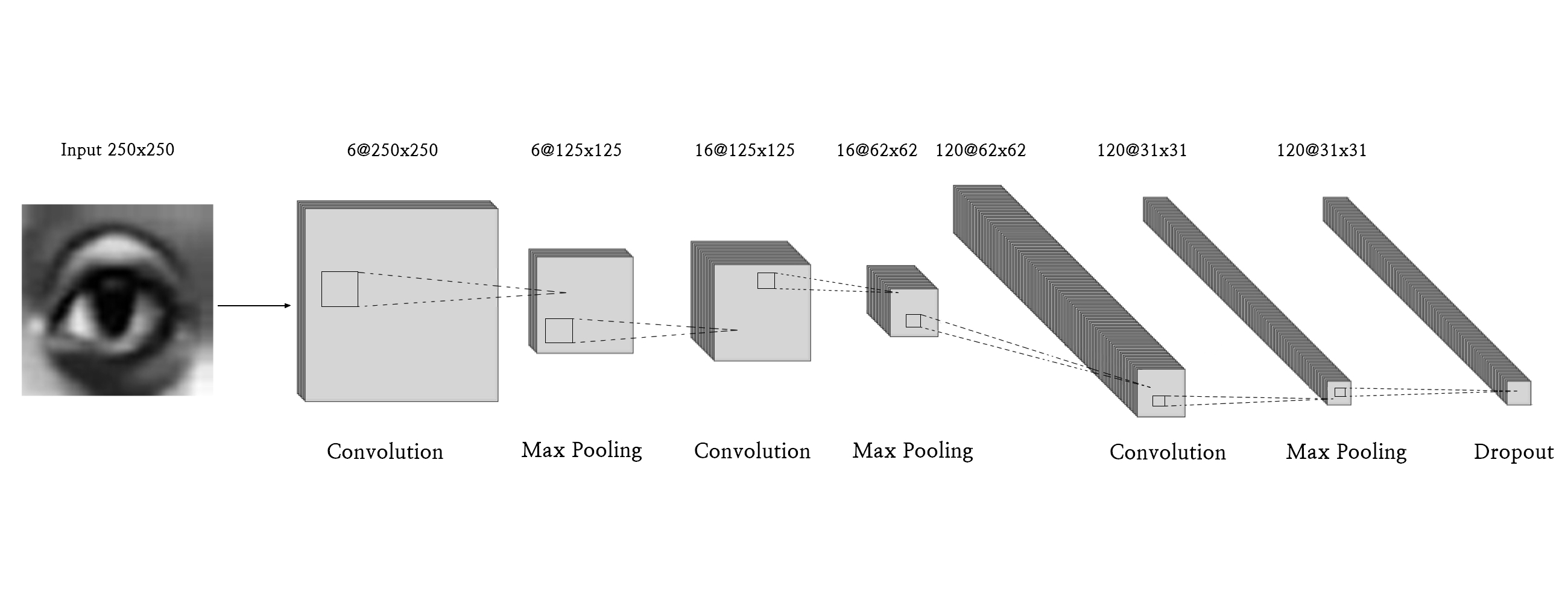}
  \caption{Structure of each sub-network for each facial patch in the proposed MFP-CNN architecture. The input of each sub-network is a facial patch and the output is fused in a concatenation layer.}
  \label{sub-network}
\end{figure*}

\subsection{Multi-Facial Patches Aggregation Convolutional Neural Networks}
\label{multi_patches_based_cnn}
The seven extracted patches from each face image are fed to seven sub-networks. 
The networks' responses are then fused and concatenated in two dense layers separated by a dropout layer. Particular features are learnt from each patch using different sub-networks and the concatenate layer aggregate all sub-networks together.  Fig.~\ref{MFP-CNN} shows the architecture of the proposed Multi-Facial Patches Agreggation Convolutional Neural Network (MFP-CNN). The sub-network for each patch is composed of three convolutional layers followed by a max pooling layer. Dropout is performed. The structure of each sub-network for each patch is shown in Fig~\ref{sub-network}. \\

\textbf{Implementation details and over-fitting problem :}  

Many parameters may affect the performance of the MFP-CNN such as the number of layers in each sub-network. 
Due to the small size of patches, there is no need to have a very deep architecture. Thus, we propose the following architecture for each of the sub-networks: 
\begin{itemize}
\item (C1) First convolutional and  max pooling layers: 6 channels. 
\item (C2) Second convolutional  and  max pooling layers: 16 channels. 
\item (C3) Third convolutional  and  max pooling layers: 120 channels. 
\end{itemize}
The filter size of the all the convolutional layers is $5\times5$ pixels and the stride of the max pooling layer is 2. 
The output of the each sub-network has $120 \times 31 \times 31=115320$ dimensions. 
Therefore, the input of the first dense layer has $ 7 \times 115320= 807240$ dimensions. 
The activation functions are ReLU for the first dense layer and Softmax for the second dense layer. 
The MFP-CNN is optimized using RMSProp optimizer with a learning rate of $10^{-3}$.

\subsection{Conditional Generative Adversarial Network for Facial Expression generation}
\label{cGAN}

The technique of artificially expanding labeled training sets by transforming data points in ways which preserve class labels known as data augmentation has quickly become a critical and effective tool to handle labeled data scarcity problem.

Generative adversarial networks (GANs) \cite{goodfellow2014generative} have been
widely used in the field of computer vision and machine learning for various applications such as image to image translation \cite{isola2017image}, face generation \cite{gauthier2014conditional}, and semantic segmentation \cite{luc2016semantic}. 
GANs learn to generate new data samples based on two 'adversarial' networks that are trained and updated at the same time.
\begin{itemize}
    \item Generator G: captures the data distribution and generates realistic samples
    \item Discriminator D: estimates the probability that a sample comes from the training data rather than G. 
\end{itemize}

To generate more facial expression images, the conditional version of the GANs is considered in this work \cite{Mirza2014}. cGANs enable the generation of fairly realistic synthetic images by forcing the generated images to be close statistically to the real ones. 

The goal of these networks is to increase the probability of samples generated by generative networks to resemble the real data so that the discriminator network fail to differentiate between the generated samples and the real data.
The generator maximizes the log-probability of labeling real and fake images correctly while the estimator minimizes it. 

From an observed image $x$ and a random noise vector $z$, cGAN learn a mapping to the output image $y$ : $G(x,y)\longrightarrow y$. This can be formulated as an optimization problem whose aim is to solve the min-max problem:

\begin{equation}
    G^{*}= \adjustlimits \argmin_{G} \max_{D} \mathcal{L}_{cGAN} (D) + \lambda \mathcal{L}_{cGAN} (G)
    \label{eq1}
\end{equation}

Where $ \mathcal{L}_{cGAN} (D)$ is the loss function of the discriminator and $ \mathcal{L}_{cGAN} (G)$ is the loss function of the generator G as defined in the following:

\begin{equation}
    \mathcal{L}_{cGAN} (G)= 1/N \sum\limits_{i=1}^n \mathcal{L}_{ad} + \alpha \mathcal{L}_{MSE} + \beta \mathcal{L}_{PEP}
    \label{eq2}
\end{equation}

where N is the total number of training images, $\mathcal{L}_{ad} $ is the adversarial loss, $\mathcal{L}_{MSE}$ and $\mathcal{L}_{PEP}$ are the pixel-wise MSE loss and the perceptual loss between the regenerated and training samples respectively as defined in \cite{YangIdentity2018}. 
The loss function of the discriminator can be expressed as :
\begin{equation}
    \mathcal{L}_{cGAN} (D)= 1/N \sum\limits_{i=1}^n \log D(x,y) + \log (1-D( x , G(x,z) ))
    \label{eq3}
\end{equation}

\begin{figure*}
	\centering
		\begin{subfigure}{0.49\textwidth} 
		\includegraphics[width=\textwidth]{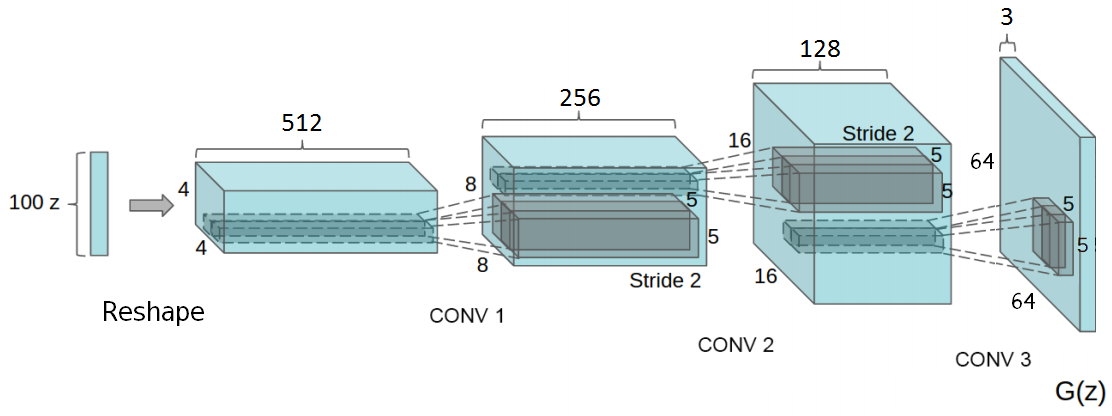}
		\caption{Generator G.} 
	\end{subfigure}
	\vspace{1em} 
	\begin{subfigure}{0.49\textwidth} 
		\includegraphics[width=\textwidth]{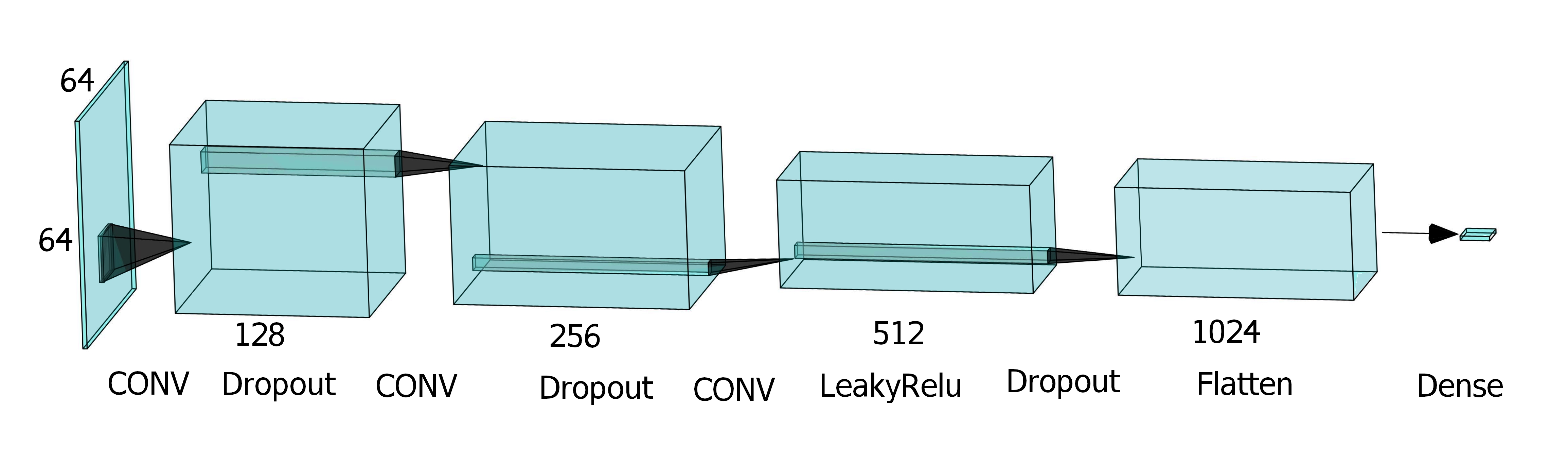}
		\caption{Discriminator D.} 
	\end{subfigure}
	\caption{Structure of the generator G and the discriminator D in cGAN used to generate new facial expression training examples.} 
	\label{discriminator_generator}
\end{figure*}

The structure of the generator G and the discriminator D are summarized in Fig.~\ref{discriminator_generator}.

\subsection{Patches generation for data augmentation}
\label{sec_patches_generation}
To augment the number of training examples further more and increase the robustness of the proposed model to random transformations, we propose a tranformation-based technique to generate facial patches.
Such technique for data augmentation leverages user-domain knowledge in the form of transformation operations. 
The problem is formulated as a generative model over a set of transformation functions (TFs) which are user-specified operators representing geometric transformations to data points. 
Five TFs are considered which rotate the facial patch by 90 and 180 degrees, translate it, shift it and transform it with Zero Component Analysis  (ZCA) whitening. 

\begin{table*}
\begin{tabular}{|l||p{5cm}|p{3cm}|p{2cm}|p{2cm}|}
\hline
\textbf{Paper} & \textbf{Method} & \textbf{Data selection} & \textbf{Expressions} & \textbf{Accuracy} \\
\hline \hline
Zhang et al. (2017) \cite{Zhang2017} & Deep Evolutional Spatial-Temporal Networks (PHRNN-MSCNN) & - & 7  & $98.50\%$ \\
\hline
Sun et al. (2017) \cite{Sun2019} & Multi-channel Deep Spatial-Temporal Feature Fusion Network (MDSTFN) & 1st frame (neutral) + seven successive (peak) & 6  & $98.38\%$ \\
\hline
Kuo et al. (2018) \cite{Kuo2018} & frame-to-sequence approach & 9 frames & 7  &  $98.47\%$ \\
\hline
Kim et al. (2017) \cite{Kim2017} & Deep Generative-Contrastive Network & - & 7  & $97.93\%$  \\
\hline
Yang et al. (2018) \cite{YangIdentity2018}& Identity-Adaptive Generation (IA-gen) & last three frames & 7  & $96.57\%$   \\
\hline
Jung et al. (2015) \cite{Jung2015} & Deep Temporal Appearance-Geometry Network (DTAGN) & - & 7  & $97.25\%$ \\
\hline
Yu et al. (2018) \cite{Yu2018} & Deeper Cascaded Peak-piloted Network (DCPN) & 7th to 9th (weak) last one to 3 frames (strong) & 7  & $98.6\%$  \\
\hline
Zhao et al. (2016) \cite{Zhao2016} & Peak-Piloted Deep Network (PPDN)& last 3 frames (peak) & 7  & $97.3\%$ \\
\hline
\hline
Our method & Multi-Facial Patches Aggregation Convolutional Neural Networks (MFP-CNN) & 1st to 3rd (neutral)+ 4th to last frame (expression)  &  8   & $98.07\%$\\
\hline
\end{tabular}
 \caption{Comparison of the proposed MFP-CNN architecture and state-of-the-art deep learning based FER methods on the CK+ database.}
  \label{tab:comparison_existing_approaches}
\end{table*}

\begin{table*}
  \centering
    \begin{tabular}{|c||l|c|c|}
      \hline 
      \textbf{Experiment} & \textbf{Training Dataset} & \textbf{Testing Dataset} & \textbf{Accuracy} \\
      \hline
       \hline
      1   & CK+ & CK+ & 89.77\%   \\
      2 & CK+ and images generated cGAN & CK+ & 96.60\% \\
      3 & CK+ and generated patches & CK+ & 97.96 \%  \\
     4 & CK+ and generated images and patches & CK+ & \textbf{98.07}\% \\
      \hline
    \end{tabular}
  \caption{Summary of MFP-CNN performance in different experiments and setups. }
  \label{tab:experimental_results}
\end{table*}

\begin{figure*}[]
  \centering \includegraphics[width=\textwidth]{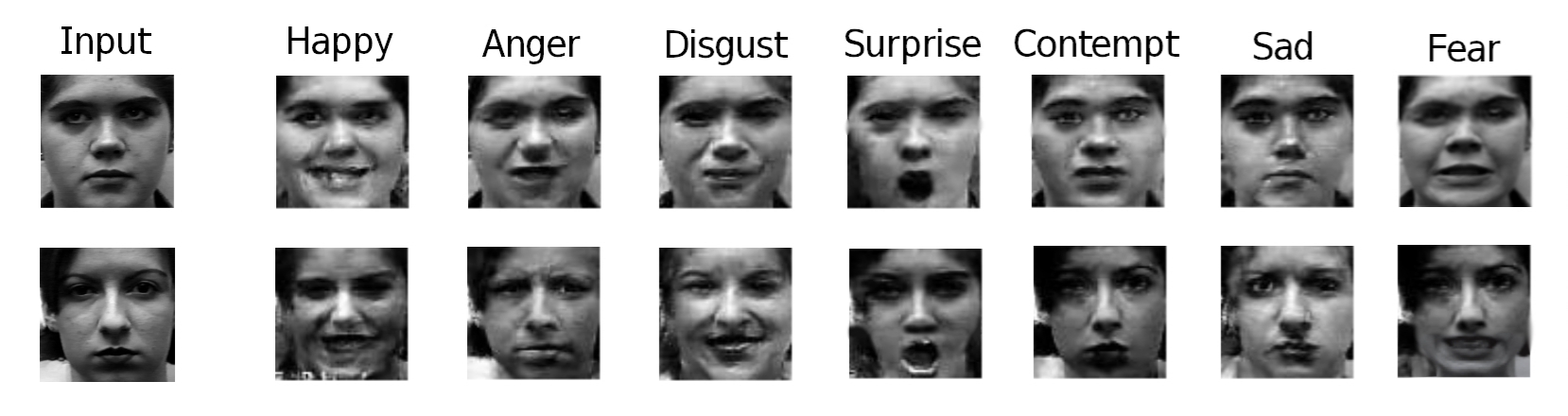}
  \caption{The generated seven facial expression images for an input image from Ck+ dataset using conditional Generative Adversarial Network  }
  \label{cGAN_results}
\end{figure*}

\begin{figure*}[]
   \centering 
   \begin{subfigure}[b]{0.49\textwidth}
         \centering
          \includegraphics[width=\textwidth]{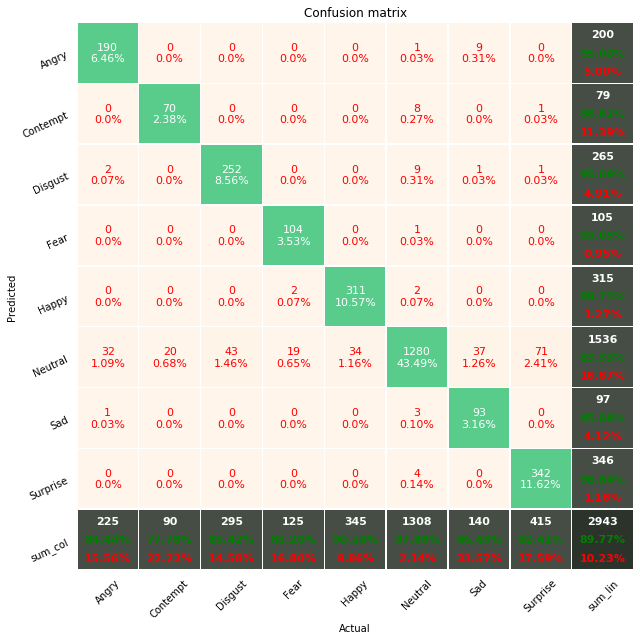}
         \caption{Experience 1.}
          \label{confusion_matrix_experience_1}
   \end{subfigure}
      \hfill  
   \centering
     \begin{subfigure}[b]{0.49\textwidth}
         \centering
          \includegraphics[width=\textwidth]{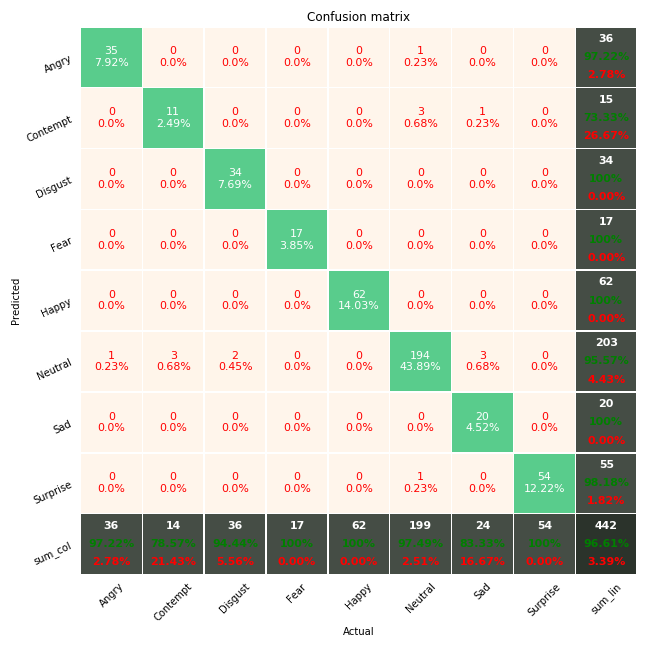}
         \caption{Experience 2.}
          \label{confusion_matrix_experience_2}
   \end{subfigure}
     \hfill 
\centering
    \begin{subfigure}[b]{0.49\textwidth}
         \centering
          \includegraphics[width=\textwidth]{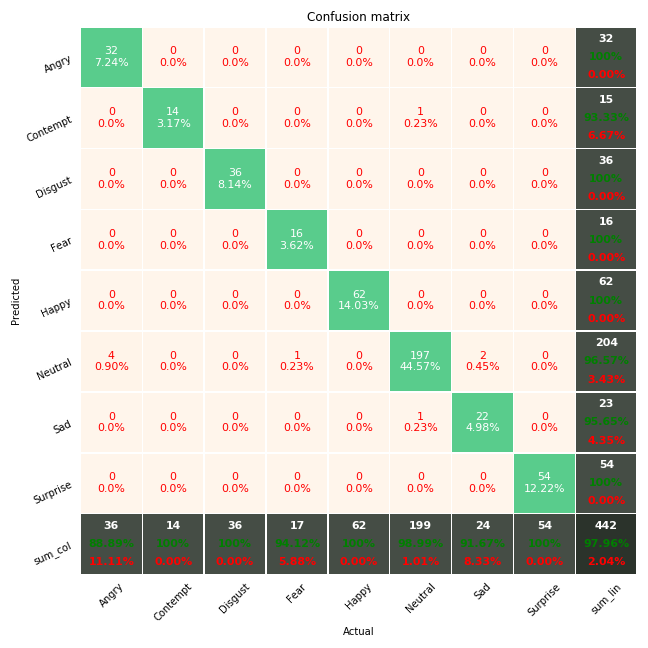}
         \caption{Experience 3.}
          \label{confusion_matrix_experience_3}
   \end{subfigure}
  \hfill 
    \centering
 \begin{subfigure}[b]{0.49\textwidth}
         \centering
          \includegraphics[width=\textwidth]{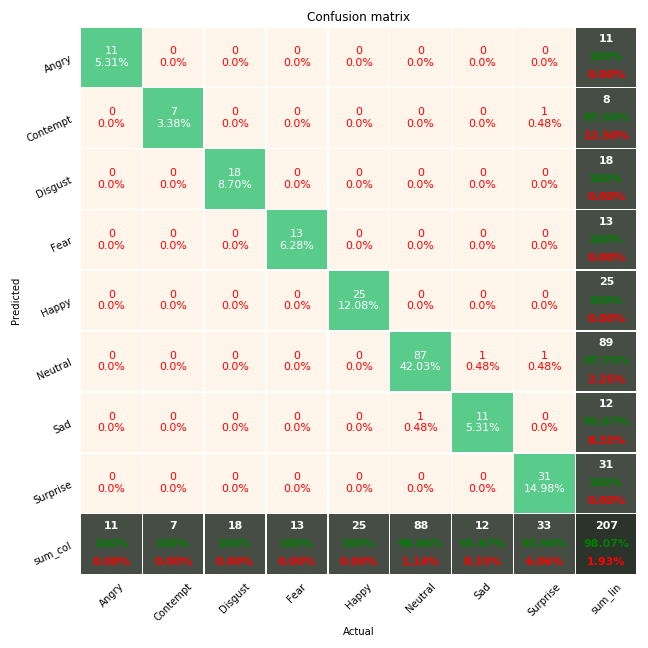}
         \caption{Experience 4.}
          \label{confusion_matrix_experience_4}
   \end{subfigure}
\caption{Confusion matrices of testing the MFP-CNN on the CK+ database of four different experiments. (a) Experience 1: training the baseline MFP-CNN on the CK+. (b) Experience 2: training on the augmented CK+ with cGAN. (c) Experience 3: training on the augmented CK+ with the transformation functions. (d) Experience 4: training on the augmented CK+ with cGAN and the transformation functions. }
\label{confusion_matrices}
\end{figure*}

 \section{Experiments and results}
 \label{results}
In this section we explain the training process of the proposed MFP-CNN and report experiments on different datasets. We used the extended Cohn-Kanade dataset (CK+) \cite{Lucey2010} as a training database to study two points: 
(1) the performance of patch-based deep learning for expression recognition,
(2) the efficiency of data augmentation for FER using cGANs and transformation functions. (3) the generalization of the proposed model to other datasets.

\subsection{Datasets}
For the experiments, the following databases are used as training and testing databases:

\begin{itemize}
    \item CK+ \cite{Lucey2010}: the extended Cohn-Kanade dataset is the main dataset used in our experiments mainly because it is one of the largest publicly available datasets for FER, which make it adequate to be considered for the development of deep learning-based approaches. 
    It is a complete dataset of Action Units and emotion-specified expressions. It includes 593 sequences from 123 subjects. The image sequence vary in duration between 10 and 60 frames and go from the onset (neutral face) to peak facial expression. 
    Frontal face images of seven discrete emotions, consisting of synchronous views of the posed expression from an angle of 30 degrees, are collected. 
    The CK+ is not balanced in term of gender and ethnicity with  $69\%$ female, $81\%$ Euro-American, $13\%$ Afro-American, and $6\%$ other groups. 
    In our experiments, all the frames per sequence are considered for training and testing. 
    We choose to assign the first seven frames with the label 'Neutral'. Moreover, the last three frames corresponding to  high-intensity expression are labeled with the facial expression label of the sequence.
\end{itemize}
 
To evaluate the generalization power of the proposed model to other datasets, two datasets are considered. 
 \begin{itemize}
     \item JAFFE \cite{Lyons1998}: the Japanese Female Facial Expression database acquired in laboratory-controlled conditions exactly like the main dataset used to train the model. Ten subjects posed six facial expressions (happiness, sadness, surprise, anger, disgust, fear) and a neutral face for a total of 219 images. Only Japanese females participate in the data collection.
     
     \item SFEW \cite{DhallVideo2015}: the Static Facial Expression in the Wild 2.0 dataset acquired in the wild conditions which could be more challenging when the dataset is trained with a laboratory-controlled dataset. 
     It was introduced during the sub-challenge of EmotiW 2015 \cite{DhallVideo2015}. 
     It contains 1766 facial expression images of seven discrete emotions. 
     It is created from AFEW 5.0 videos dataset \cite{dhall2011acted}.
     A semi-automatic technique based on k-means clustering was performed for the data labelling.

 \end{itemize}

\subsection{Experimental results}
In order to access the performance of the proposed MFP-CNN-based approach for FER, we perform several experiments  as shown in Table~\ref{tab:experimental_results}:
\begin{itemize}
    \item  MFP-CNN framework evaluation :train and test on CK+ dataset with ten fold subject-independent cross-validation protocol. We divide the subjects frames into 10 subsets. We avoid subjects appearing in both the train and the test folds simultaneously. Four experiments are realized: 
    \begin{itemize}
        \item [][Experiment \#1] : Without data augmentation 
        \item [][Experiment \#2] : With cGAN data augmentation 
        \item [][Experiment \#3] : With TF based data augmentation
        \item [][Experiment \#4] : With cGAN and TF data augmentation
    \end{itemize}
    \item Generalization power of the MFP-CNN model:
    \begin{itemize}
    \item Train on CK+ and test on JAFFE
    \item Train on CK+ and test on SFEW
    \end{itemize}
\end{itemize}

\subsubsection{MFP-CNN framework evaluations }
\label{MFP-CNN-framework-evaluations}
  
 \textbf{Evaluation of the baseline MFP-CNN on the CK+ : }
 In the first experiment, we assess the performance of the baseline MFP-CNN (without data augmentation) on the well-known benchmark of the CK+. We obtain an overall accuracy of 89.77\%. 
 This result shows that the baseline MFP-CNN is robust in learning facial expressions on a relatively small dataset. 
 Even without expanding the labeled training sets or using transfer learning \cite{Yu2015,NgDeep2015}, the accuracy is sufficiently acceptable.
 The corresponding confusion matrix is shown in Fig.~\ref{confusion_matrix_experience_1}. 
As we see from the figure, the mis-classified face images correspond mainly to neutral faces. 
This can be justified by the fact that the neutral face images in the CK+ dataset correspond to the first seven frames of the image sequence. These images go from the neutral face to the peak face expression. Looking at these frames, we notice that starting from the third frame, the images labeled as neutral begin to be close in term of intensity to the peak facial expression (7th frame).  
This influences the dynamic variation of key parts of the face causing mis-classification of some neutral faces as the class of the peak face expression. \\
 
 \textbf{Evaluation of data augmentation: } 
 In the second experience, we evaluate the performance of generating more facial images using cGAN (section~\ref{cGAN}). As shown in Figure~\ref{cGAN_results}, the generated facial images are fairly realistic, the different facial expressions can be identified visually by a human while the identity-related information is preserved. 
 The regenerated images using cGAN are used with the original images of the CK+ to train the MFP-CNN. 
 This second experiment outperforms the first with an accuracy of 96.60\% and shows that data augmentation improves the accuracy of the proposed MFP-CNN-based approach. 
 The corresponding confusion matrix is shown in Fig.~\ref{confusion_matrix_experience_2}. 
 The most mis-classified class is `Contempt' with $73.33\%$ good classification. The $26.67\%$ of mis-classified face images are confused with neutral or sad face images. 
 This can be explained by the fact that these three expressions (contempt, sad, neutral) can be very similar in appearance for some people.  Thus there is not a large difference between these facial expressions in term of the dynamic variation of key parts of the human face.
 
 In the third experiment, we evaluate the performance of augmenting training data by generating more patches to train the MFP-CNN (section ~\ref{sec_patches_generation}).
 We obtain an accuracy of $97.96\%$. Compared to the baseline framework of MFP-CNN in the first experience, the accuracy was increased by $8.25\%$.
This confirms the efficiency of augmenting pre-training datasets in this framework of FER. 
 The corresponding confusion matrix is shown in Fig.~\ref{confusion_matrix_experience_3}. Only three facial expression classes are mis-classified: neutral, sad and contempt. Compared to using cGAN for data augmentation, the mis-classifition rate of these expressions was significantly reduced.
 
 In the fourth experiment, the two data augmentation techniques are considered and more facial images and facial patches are generated using cGAN and the transformation functions, respectively. Using both data augmentation techniques, we obtain an accuracy of $98.07\%$, which is slightly better than using only transformation functions. 
 The corresponding confusion matrix is shown in Fig.~\ref{confusion_matrix_experience_4}. The mis-classified samples also belong to sad, neutral and contempt. 
 We can conclude that data augmentation can remarkably increase the efficiency of deep learning based approaches, without the need for acquiring new data or applying transfer learning techniques.
 

\subsubsection{Generalization of the proposed model to other datasets}
FER datasets present a big variability due to the variations in facial expressions among different persons, as well as the variability in  data labeling and conditions for data acquisition. 
A robust model learns from known examples and has the capacity to generalize to new examples acquired under different conditions. 
In the following, we evaluate the generalization power of the proposed model for dataset bias. Knowing that CK+ is a standard dataset acquired in laboratory-controlled conditions where face images are taken in same conditions, we perform two experiments. We train the MFP-CNN on CK+ and test on two different datasets: one on a dataset with similar laboratory-controlled conditions (JAFFE) and the other on a dataset acquired in a more challenging in the wild conditions (SFEW).

\begin{figure}[]
	\begin{subfigure}[b]{\columnwidth} 
		\scalebox{0.9}{\includegraphics[width=\textwidth]{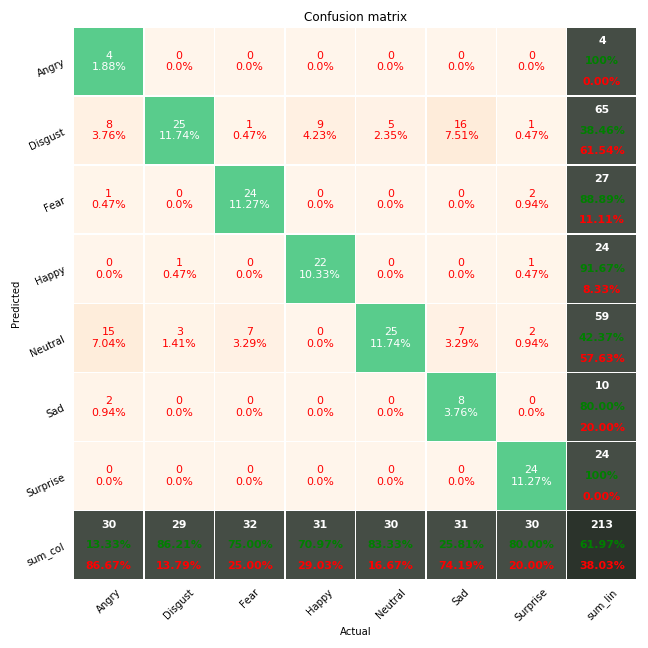}}
		\caption{Training on CK+ and testing on JAFFE} 
		\label{confusion_matrix_jaffe}
	\end{subfigure}
	\begin{subfigure}[b]{\columnwidth} 
		\scalebox{0.9}{\includegraphics[width=1\textwidth]{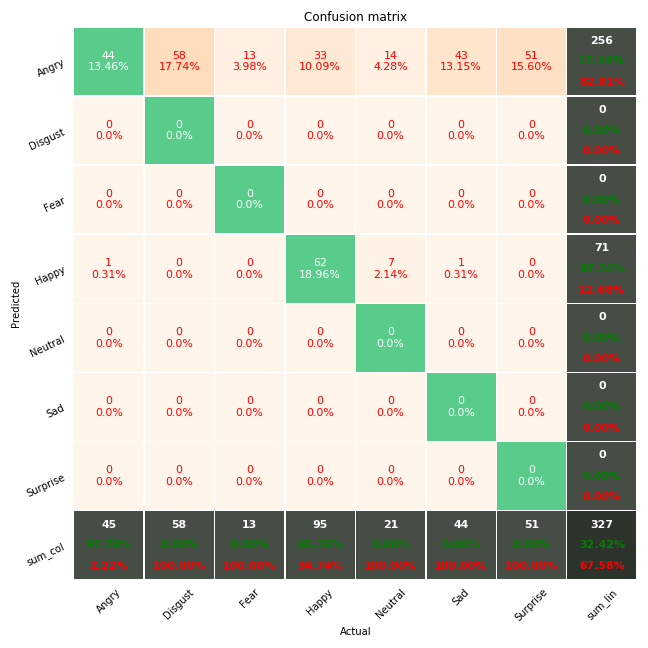}}
		\caption{Training on CK+ and testing on SFEW} 
		\label{confusion_matrix_afew}
	\end{subfigure}
	\begin{subfigure}[b]{\columnwidth} 
		\scalebox{0.9}{\includegraphics[width=1\textwidth]{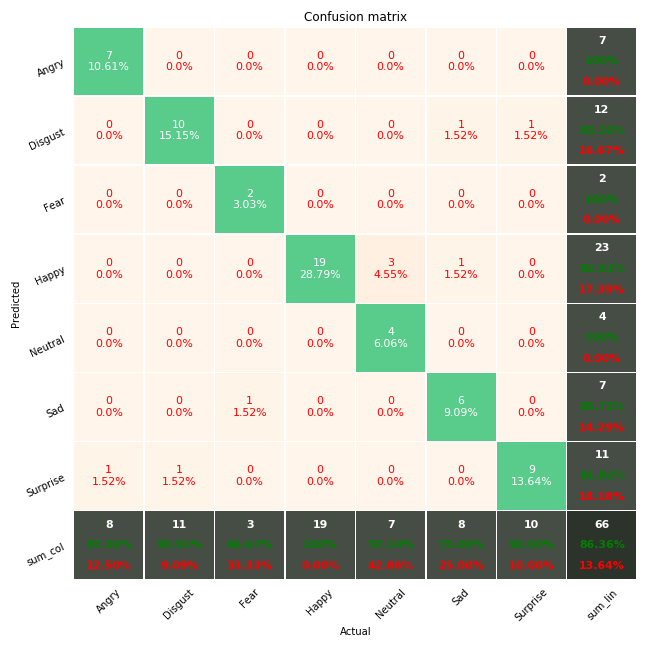}}
		\caption{Training on CK+, fine-tuning on 80\% of SFEW and testing on 20\% of SFEW} 
		\label{confusion_matrix_afew_20}
	\end{subfigure}
	\caption{Confusion matrices of the MFP-CNN trained the CK+ database and tested on others FER databases.} 
	\label{confusion_matrices_generalization}
\end{figure}

\begin{itemize}
   \item \textbf{Training on CK+ and testing on JAFFE:} 
   in the first experiment, the generalization of model to a dataset acquired in laboratory-controlled conditions is evaluated. 
   The test dataset presents same gender and same ethnicity, unlike the train dataset which contains different ethnicities and two genders. An accuracy of $61.97\%$ is achieved and the corresponding confusion matrix is shown in Fig.~\ref{confusion_matrix_jaffe}. The misclassifcation comes mainly from two facial expressions: disgust and neutral. $61.54\%$ of disgust face images and $51.63\%$ of neutral face images are misclassified and they are confused with all facial expressions present in JAFFE dataset. Despite the big difference between the JAFFE and the CK+ databases in term of image resolution, number of face expressions, conditions of acquisition, gender and ethnicity, the learned MFP-CNN on CK+ is still able to predict correctly the facial expression in $62\%$ of cases. 
   It is a promising result.
    
    \item \textbf{Training on CK+ and testing on SFEW:} in the second experiment, the test dataset is collected in the wild for unconstrained FER. Training on laboratory-controlled dataset and testing on labeled facial expressions in the wild dataset is more challenging. Training on CK+ and testing on SFEW gives an accuracy of only $32.7\%$ as shown in confusion matrix of Figure~\ref{confusion_matrix_afew}. When fine-tuning is applied, the accuracy of FER increases considerably and achieves $86.36\%$ as shown in the confusion matrix of Figure~\ref{confusion_matrix_afew_20}. The transition from ideal conditions of face acquisition (CK+) to faces in the wild conditions (SFEW) shows an increased error rate, a fine-tuning improved considerably the results.
\end{itemize}

\subsubsection{Comparison with existing methods}
 In this section, we compare the performance of the proposed method with state-of-the-art FER methods on the CK+ dataset. Only deep neural networks-based architectures are reported in this comparison as they perform better than shallow learning-based methods on FER. The top performing methods in FER literature exceed $98\%$ \cite{Yu2018,Zhang2017,Sun2019,Kuo2018}. 

 Two factors that could impact strongly the performances of FER methods on CK+ database, are not standard in the existing FER methods:
 \begin{itemize}
     \item Data selection: the number of frames to consider , 
     \item Sequence-based methods outperform frame-based methods  \cite{Kuo2018,Yu2018}.
 \end{itemize}
 In \cite{Kuo2018}, the high performance is achieved thanks to the extension of the proposed framework to a frame-to-sequence approach by exploiting the temporal information with gated recurrent units. The frame-based approach achieve only $97.37\%$. The use of peak expression to supervise the non-peak expression improves the performance of the proposed framework in \cite{Yu2018}. Also learning spatio-temporal features that capture the dynamic variation of facial physical structure and the dynamic temporal evolution of expression \cite{Zhang2017,Sun2019} boosts the performance of FER methods. 
 In \cite{Sun2019}, only 6 universal emotional faces are tested. Neutral and contempt facial expressions are ignored. We have reported in section~\ref{MFP-CNN-framework-evaluations} that the mis-classified samples belong mainly to these two classes. Also, they do not use three consecutive face images to reduce the sample correlation caused by two similar face images. 
 This can explain the achieved high good performance. \\
 Our frame-based approach achieves an accuracy of $98.07\%$ on 8 facial expressions. To the best of our knowledge, it is the only approach reported in the FER literature that consider neutral facial expressions despite its correlation with the rest of the seven facial expressions on the CK+ database. 
 Despite the correlation between the third frame labeled in our experiments as neutral and the fourth frame labeled as a facial expression, the MFP-CNN framework performs well among existing FER methods.

\section{Conclusion and future works}
 \label{conclusion}
In this paper, we proposed a Multi-Facial Patches Aggregation Convolutional Neural Networks (MFP-CNN) for Face Expression Recognition. Our approach combines patch-based expression recognition approaches with deep learning. Further, we propose to expand the labeled training by generating facial expressions images and patches. Two techniques are considered for that by using the Conditional Generative Adversarial Networks and a set of transformation functions. The proposed method achieves significantly good performance in comparison to other existing approaches on the CK+ database. The performance decreases significantly when tested for database bias. For future work, we plan to investigate how we can account for database bias and make the model generalize well to new databases acquired in different conditions and in real-world scenarios.



\bibliography{main}
\bibliographystyle{spmpsci}      

\end{document}